\documentclass[10pt]{article}
\usepackage[utf8]{inputenc}
\usepackage[margin=1in]{geometry}
\usepackage{rotating}
\usepackage{booktabs}
\usepackage{array}
\usepackage{amsmath, amsfonts, amssymb}
\usepackage[table,xcdraw]{xcolor}
\usepackage{hyperref}
\usepackage{bbm}
\usepackage{algorithm}
\usepackage[noend]{algpseudocode}
\usepackage{subcaption}
\usepackage{graphbox}
\usepackage{accents}

\usepackage{hhline}
\usepackage{colortbl}

\usepackage{tabularx,booktabs}
\newcolumntype{Y}{>{\centering\arraybackslash}X}

\usepackage{makecell}

\usepackage[numbers,sort]{natbib}

\usepackage{mathtools}
\mathtoolsset{showonlyrefs}

\usepackage{amsthm}
\usepackage[normalem]{ulem}
\usepackage{soul}

\usepackage{chngcntr}
\usepackage{apptools}
\AtAppendix{\counterwithin{theorem}{section}}
\AtAppendix{\counterwithin{prop}{section}}
\AtAppendix{\counterwithin{cor}{section}}
\AtAppendix{\counterwithin{lemma}{section}}

\usepackage{mathtools}

\usepackage{multicol}
\usepackage{multirow}

\definecolor{Gray}{gray}{0.8}
\definecolor{LightGray}{gray}{0.95}

\def\R{\mathbb{R}}
\def\E{\mathbb{E}}

\def\N{\mathbb{N}}

\renewcommand{\P}{\mathbb{P}}

\newcommand{\ppi}{\texttt{PPI}}
\newcommand{\ppboot}{\texttt{PPBoot}}

\newcommand{\crossppi}{\texttt{cross-PPI}}
\newcommand{\crossppboot}{\texttt{cross-PPBoot}}

\newcommand{\Var}{\mathrm{Var}}
\newcommand{\Cov}{\mathrm{Cov}}

\newcommand{\Xt}{\tilde{X}}

\usepackage{esvect}

\makeatletter
\def\blfootnote{\xdef\@thefnmark{}\@footnotetext}
\makeatother

\title{A Note on the Prediction-Powered Bootstrap}

\author{Tijana Zrnic\\ 
Department of Statistics and Stanford Data Science\\
Stanford University}
\date{}

\begin{document}

\maketitle

\begin{abstract}
We introduce \texttt{PPBoot}: a bootstrap-based method for prediction-powered inference. \texttt{PPBoot} is applicable to arbitrary estimation problems and is very simple to implement, essentially only requiring one application of the bootstrap.
Through a series of examples, we demonstrate that \texttt{PPBoot} often performs nearly identically to (and sometimes better than) the earlier \texttt{PPI(++)} method based on asymptotic normality---when the latter is applicable---without requiring any asymptotic characterizations. Given its versatility, \ppboot~could simplify and expand the scope of application of prediction-powered inference to problems where central limit theorems are hard to prove.
\end{abstract}

\section{Introduction}

Black-box predictive models are increasingly used to generate efficient substitutes for gold-standard labels when the latter are difficult to come by. For example, predictions of protein structures are used as efficient substitutes for slow and expensive experimental measurements \cite{jumper2021highly, bludau2022structural, barrio2023clustering}, and large language models are used to cheaply generate substitutes for scarce human annotations \cite{zheng2024judging, boyeau2024autoeval, fan2024narratives}. Prediction-powered inference (\ppi) \cite{angelopoulos2023prediction} is a recent framework for statistical inference that combines a large amount of machine-learning predictions with a small amount of real data to ensure simultaneously valid and statistically powerful conclusions.

While \ppi~\cite{angelopoulos2023prediction} (and its improvement \texttt{PPI++} \cite{angelopoulos2023ppipp}) offers a principled solution to incorporating black-box predictions into the scientific workflow, its scope of application is still limited. The current analyses focus on certain convex M-estimators such as means, quantiles, and GLMs to ensure tractable implementation. Furthermore, applying \ppi~requires case-by-case reasoning: inference relies on a central limit theorem and problem-specific plug-in estimates of the asymptotic variance. This makes it difficult for practitioners to apply \ppi~to entirely new estimation problems.

We introduce \ppboot: a bootstrap-based method for prediction-powered inference, which is applicable to arbitrary estimation problems and is very simple to implement. \ppboot~does not require any problem-specific derivations or assumptions such as convexity. Across a range of practical examples, we show that \ppboot~is valid and typically at least as powerful as the earlier \ppi~\cite{angelopoulos2023prediction} and \texttt{PPI++}~\cite{angelopoulos2023ppipp} methods. We also develop two extensions of \ppboot: one incorporates power tuning \cite{angelopoulos2023ppipp}, improving the power of basic \ppboot; the other incorporates cross-fitting \cite{zrnic2023cross} when a good pre-trained model for producing the predictions is not available a priori but needs to be trained or fine-tuned. Overall, \ppboot~offers a simple and versatile approach to prediction-powered inference.


Our approach to debiasing predictions is inspired by \ppi\texttt{(++)} \cite{angelopoulos2023prediction, angelopoulos2023ppipp}, but differs in that our confidence intervals are based on bootstrap simulations, rather than a central limit theorem with a plug-in variance estimate. Furthermore, our approach enjoys broad applicability, going beyond convex M-estimators.
A predecessor of \ppi, called post-prediction inference (\texttt{postpi}) \cite{wang2020methods}, was motivated by inference problems in a similar setting, with little gold-standard data and abundant machine-learning predictions. Like our method, the \texttt{postpi} method also leverages the bootstrap. However, \ppboot~is quite different and has provable guarantees for a broad family of estimation problems. 

\ppboot~is implemented in the \href{https://github.com/aangelopoulos/ppi_py}{\texttt{ppi\_py} package}.

\paragraph{Problem setup.}
We have access to $n$ labeled data points $(X_i, Y_i),~i\in[n]$, drawn i.i.d. from $\P = \P_X \times \P_{Y|X}$, and $N$ unlabeled data points $\Xt_i,~i\in[N]$, drawn i.i.d. from the same feature distribution $\P_X$. The labeled and unlabeled data are independent. For now, we also assume that we have a pre-trained machine learning model $f$ that maps features to outcomes; we extend \texttt{PPBoot} beyond this assumption in Section \ref{sec:extensions}. Thus, $f(X_i)$ and $f(\Xt_i)$ denote the predictions of the model on the labeled and the unlabeled data points, respectively. Furthermore, we use $(X,Y)$ as short-hand notation for the whole labeled dataset, i.e. $X = (X_1,\dots,X_n)$ and $Y=(Y_1,\dots,Y_n)$; similarly, $f(X) = (f(X_1),\dots, f(X_n))$. We use $\tilde X, f(\tilde X)$, etc analogously.

Our goal is to compute a confidence interval for a population-level quantity of interest $\theta_0$. For example, we might be interested in the average outcome, $\theta_0 = \E[Y_i]$, a regression coefficient obtained by regressing $Y$ on $X$, or the correlation coefficient between a particular feature and the outcome. We use $\hat\theta(\cdot)$ to denote any ``standard'' (meaning, not prediction-powered) estimator for $\theta_0$ that takes as input a labeled dataset. In other words, $\hat\theta(X,Y)$ is any standard estimate of $\theta_0$. For example, if $\theta_0$ is a mean over the data distribution, $\theta_0=\E[g(X_i,Y_i)]$ for some $g$, then $\hat\theta$ could be the corresponding sample mean: $\hat\theta(X,Y) = \frac{1}{n} \sum_{i=1}^n g(X_i, Y_i)$. Unlike existing \texttt{PPI} methods, which focused on M-estimation, \texttt{PPBoot} does not place restrictions on $\theta_0$ and can be applied as long as there is a sensible estimator $\hat\theta$.

\section{PPBoot}

We present a bootstrap-based approach to prediction-powered inference that is applicable to arbitrary estimation problems. 
The idea is very simple. Let $B$ denote a user-chosen number of bootstrap iterations. At every step $b\in[B]$, we resample the labeled and unlabeled data with replacement; let $(X^*,Y^*)$ and $\tilde X^*$ denote the resampled datasets. We then compute the bootstrap estimate for iteration $b$ as
$$\theta_b^* = \hat \theta(\Xt^*,f(\Xt^*)) + \hat \theta(X^*,Y^*) - \hat \theta(X^*,f(X^*)),$$
where $\hat\theta$ is any standard estimator for the quantity of interest.
Finally, we apply the percentile method to obtain the \texttt{PPBoot} confidence interval:
$$\mathcal{C}^{\mathrm{\texttt{PPBoot}}} = \Big(\texttt{quantile}\left(\{\theta_b^*\}_{b=1}^B; \alpha/2 \right), \texttt{quantile}\left(\{\theta_b^*\}_{b=1}^B; 1-\alpha/2 \right) \Big),$$
where $\alpha$ is the desired error level. We summarize \texttt{PPBoot} in Algorithm \ref{alg:ppboot}.

\begin{algorithm}[b]
\caption{\texttt{PPBoot}}
\label{alg:ppboot}
\begin{algorithmic}[1]
\Require labeled data $(X,Y)$, unlabeled data $\tilde X$, model $f$, error level $\alpha\in(0,1)$, bootstrap iterations $B\in\N$
\For{$b = 1,\dots,B$}
\State Resample $(X^*,Y^*)$ and $\tilde X^*$ from $(X,Y)$ and $\tilde X$ with replacement
\State Compute $\theta_b^* = \hat \theta(\Xt^*,f(\Xt^*)) + \hat \theta(X^*,Y^*) - \hat \theta(X^*,f(X^*))$
\EndFor
\Ensure $\mathcal{C}^{\mathrm{\texttt{PPBoot}}} = (\texttt{quantile}\left(\{\theta_b^*\}_{b=1}^B; \alpha/2 \right), \texttt{quantile}\left(\{\theta_b^*\}_{b=1}^B; 1-\alpha/2 \right) )$
\end{algorithmic}
\end{algorithm}

The validity of $\mathcal{C}^{\mathrm{\texttt{PPBoot}}}$ follows from the standard validity of the bootstrap. The key observation is that $\theta^*_b$ is a consistent estimate if $\hat\theta$ is a consistent estimator: indeed, $\hat \theta(X^*, Y^*)$ converges to $\theta_0$ and $\hat \theta(\Xt^*,f(\Xt^*)) - \hat \theta(X^*,f(X^*))$ simply estimates zero due to the fact that the labeled and unlabeled data follow the same distribution. Furthermore, not only is our estimation strategy consistent, but it is also asymptotically normal around $\theta_0$ when $\hat\theta(X,Y)$ yields asymptotically normal estimates (such as in the case of M-estimation), under only mild additional regularity. For mathematical details, we refer the reader to the work of \citet{yang2019combining}, who propose a similar estimator for average causal effects relying on this fact. The asymptotic normality implies that it would also be valid to compute CLT intervals centered at $\hat\theta^{\ppboot} = \hat \theta(\Xt,f(\Xt)) + \hat \theta(X,Y) - \hat \theta(X,f(X))$ with a bootstrap estimate of the standard error via $\theta^*_b$, though we opted for the percentile bootstrap for conceptual simplicity. Many other variants of \ppboot~based on other forms of the bootstrap are possible; see \citet{efron1994introduction} for other options.

Furthermore, we expect $\theta^*_b$ to be more accurate than the classical estimate $\hat\theta(X,Y)$ if the the machine-learning predictions are reasonably accurate, since an accurate model $f$ yields $\hat\theta(X^*, f(X^*)) \approx \hat\theta(X^*, Y^*)$, and thus the bootstrap estimate is roughly $\theta^*_b \approx \hat \theta(\Xt^*,f(\Xt^*))$. Since $\hat\theta(\Xt^*,f(\Xt^*))$ leverages $N\gg n$ data points, we expect it to have far lower variability than $\hat \theta(X,Y)$.


\section{Applications}

We evaluate \ppboot~in a series of applications, comparing it to the earlier \texttt{PPI} and \texttt{PPI++} methods \cite{angelopoulos2023prediction, angelopoulos2023ppipp}, as well as ``classical'' inference, which uses the labeled data only. To avoid cherry-picking example applications, we primarily focus on the datasets and estimation problems studied by \citet{angelopoulos2023prediction}. To showcase the versatility of our method, we run additional experiments with estimation problems not easily handled by \ppi. For now, we do not use power tuning in \texttt{PPI++}; we will return to power tuning in the next section.

Over $100$ trials, we randomly split the data (which is fully labeled) into a labeled component of size $n$ and treat the remainder as unlabeled. The confidence intervals are computed at error level $\alpha=0.1$. We report the interval width and coverage averaged over the $100$ trials, for varying $n$. To compute coverage, we take the value of the quantity of interest on the whole dataset as the ground truth. We also plot the interval computed by each method for three randomly chosen trials, for a fixed $n$. We apply \ppboot~with $B=1000$.

Each of the following applications defines a unique estimation problem on a unique dataset. We briefly describe each application; for further details, we refer the reader to \cite{angelopoulos2023prediction}.

\begin{figure}[b]
\centering
\includegraphics[width=0.93\textwidth]{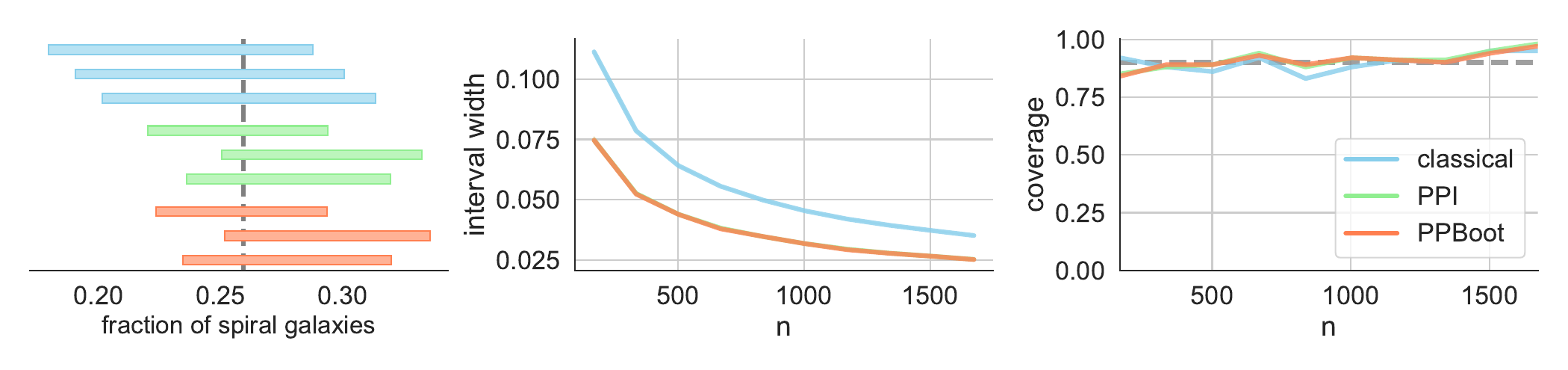}
\caption{Classical inference, \ppi, and \ppboot, applied to estimating the fraction of spiral galaxies from galaxy images.}
\label{fig:galaxy}
\end{figure}

\paragraph{Galaxies.} In the first application, we study galaxy data from the Galaxy Zoo 2 dataset \cite{willett2013galaxy}, consisting of human-annotated images of galaxies from the Sloan Digital Sky 
Survey \cite{york2000sloan}.
The quantity of interest is the fraction of spiral galaxies, i.e., the mean of a binary indicator $Y_i\in\{0,1\}$ which encodes whether a galaxy has spiral arms. We use the predictions by \citet{angelopoulos2023prediction}, which are obtained by fine-tuning a pre-trained ResNet on a separate subset of galaxy images from the Galaxy Zoo 2.
We show the results in Figure \ref{fig:galaxy}. We observe that \texttt{PPI} and general \texttt{PPBoot} have essentially identical interval widths, significantly outperforming classical inference based on a standard CLT interval. All methods approximately achieve the nominal coverage.

\begin{figure}[t]
\centering
\includegraphics[width=0.93\textwidth]{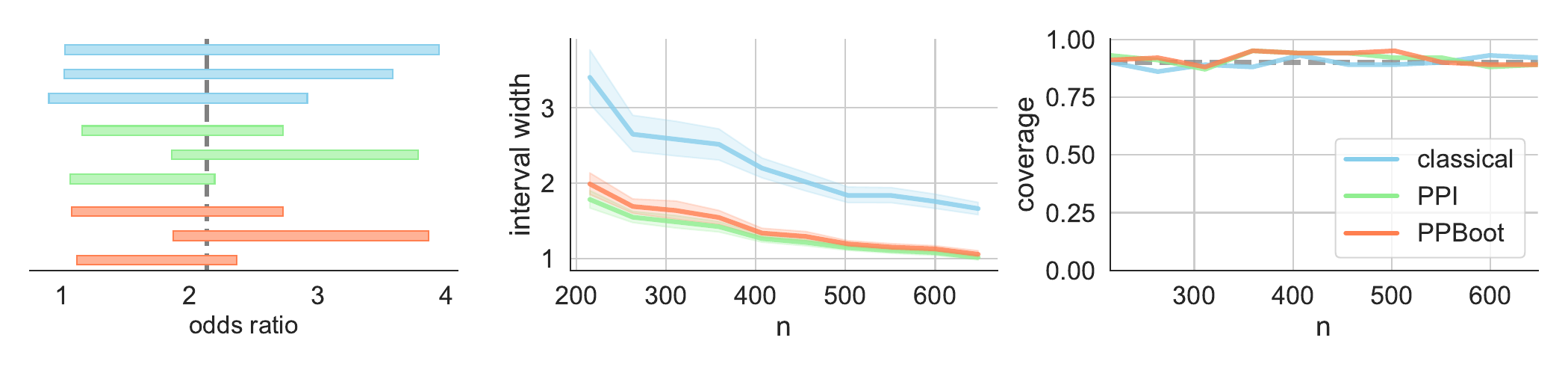}
\caption{Classical inference, \ppi, and \ppboot, applied to estimating the odds ratio between protein phosphorylation and protein disorder with AlphaFold predictions.}
\label{fig:alphafold}
\end{figure}

\paragraph{AlphaFold.} The next example concerns estimating a particular odds ratio between two binary variables: phosphorylation, a regulatory property of a protein, and disorder, a structural property. This problem was studied by \citet{bludau2022structural}. Since disorder is difficult to measure experimentally, AlphaFold \cite{jumper2021highly} predictions are used to impute the missing values of disorder. For the application of \ppi, we apply the asymptotic analysis based on the delta method provided in \cite{angelopoulos2023ppipp}, as it is more powerful than the original analysis in \cite{angelopoulos2023prediction}.
Figure \ref{fig:alphafold} shows the performance of the methods. General \texttt{PPBoot} performs similarly to \texttt{PPI} in terms of interval width, though slightly worse. As expected, classical inference based on the CLT yields much larger intervals than the other baselines. All methods achieve the desired coverage.

\begin{figure}[t]
\centering
\includegraphics[width=0.93\textwidth]{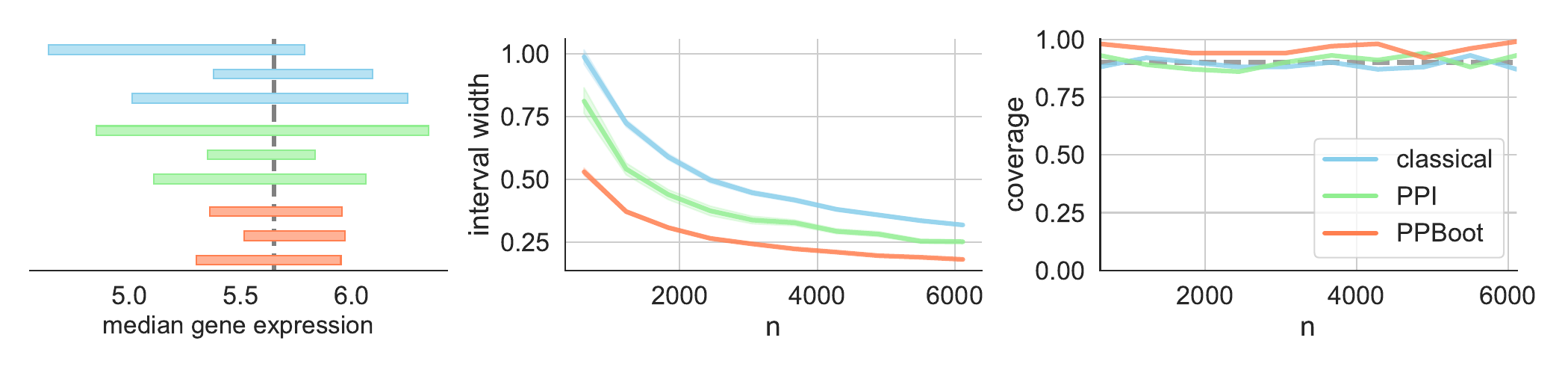}
\caption{Classical inference, \ppi, and \ppboot, applied to estimating the median gene expression with transformer predictions.}
\label{fig:gene}
\end{figure}

\paragraph{Gene expression.} Building on the analysis of \citet{vaishnav2022evolution}, who trained a state-of-the-art transformer model to predict the expression level of a particular gene induced by a promoter sequence, \citet{angelopoulos2023prediction} computed \texttt{PPI} confidence intervals on quantiles that characterize how a population of promoter sequences affects gene expression. They computed the $q$-quantile of gene expression, for $q\in\{0.25, 0.5, 0.75\}$. We report the results for estimating the median gene expression with \texttt{PPBoot}, though our findings are not substantially different for $q\in\{0.25,0.75\}$.
See Figure \ref{fig:gene} for the results. We observe that \texttt{PPBoot} leads to substantially tighter intervals than \texttt{PPI}: \texttt{PPBoot} improves over \texttt{PPI} roughly as much as \texttt{PPI} improves over classical CLT inference, all the while maintaining correct coverage.

\begin{figure}[b]
\centering
\includegraphics[width=0.93\textwidth]{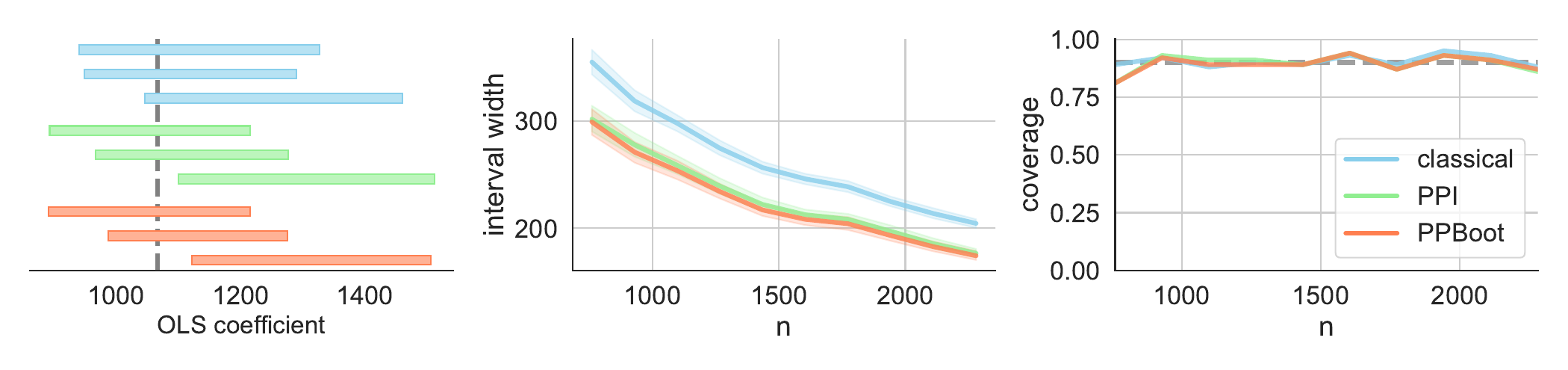}
\caption{Classical inference, \ppi, and \ppboot, applied to estimating the relationship between age and income in US census data.}
\label{fig:census_ols}
\end{figure}

\begin{figure}[t]
\centering
\includegraphics[width=0.93\textwidth]{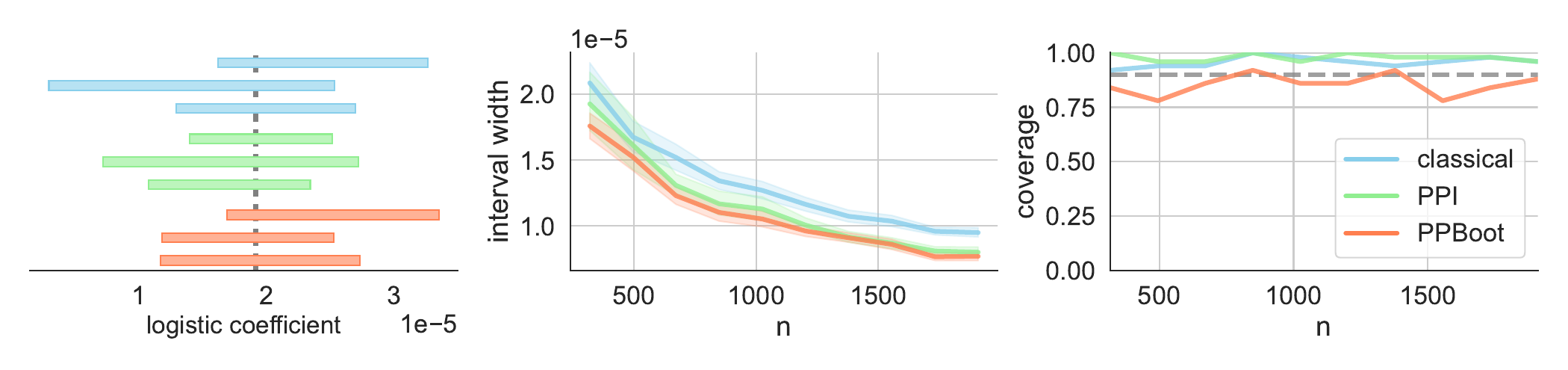}
\caption{Classical inference, \ppi, and \ppboot, applied to estimating the relationship between income and having health insurance in US census data.}
\label{fig:census_logistic}
\end{figure}

\paragraph{Census.} We investigate the relationship between socioeconomic variables in US census data, in particular the American Community Survey Public Use Microdata Sample (ACS PUMS) collected in California in 2019.
We study two applications: in the first we evaluate the relationship between age and income, and in the second we evaluate the relationship between income and having health insurance.
We use the predictions of income and health insurance, respectively, from \cite{angelopoulos2023prediction}, obtained by training a gradient-boosted tree on historical census data including various demographic covariates, such as sex, age, education, disability status, and more.
In the first application, the target of inference is the ordinary least-squares (OLS) coefficient between age and income, controlling for sex. In the second application, the target of inference is the logistic regression coefficient between income and the indicator of having health insurance. 
We plot the results for linear and logistic regression in Figure \ref{fig:census_ols} and  Figure \ref{fig:census_logistic}, respectively. In both applications, \texttt{PPBoot} yields similar intervals to \texttt{PPI}, and both methods give smaller intervals than classical inference based on the CLT. In the second application, \ppboot~slightly undercovers, though that may be resolved by increasing $B$.

\begin{figure}[t]
\centering
\includegraphics[width=0.93\textwidth]{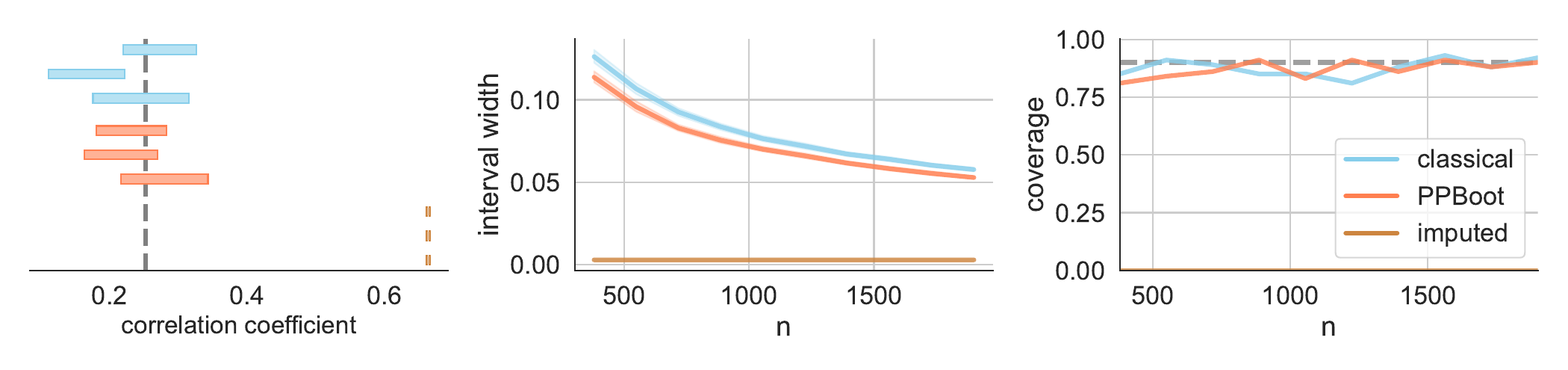}
\caption{Classical inference, \ppboot, and the imputed approach, applied to estimating the correlation coefficient between age and income in US census data.}
\label{fig:census_income_corr}
\end{figure}

\begin{figure}[b]
\centering
\includegraphics[width=0.93\textwidth]{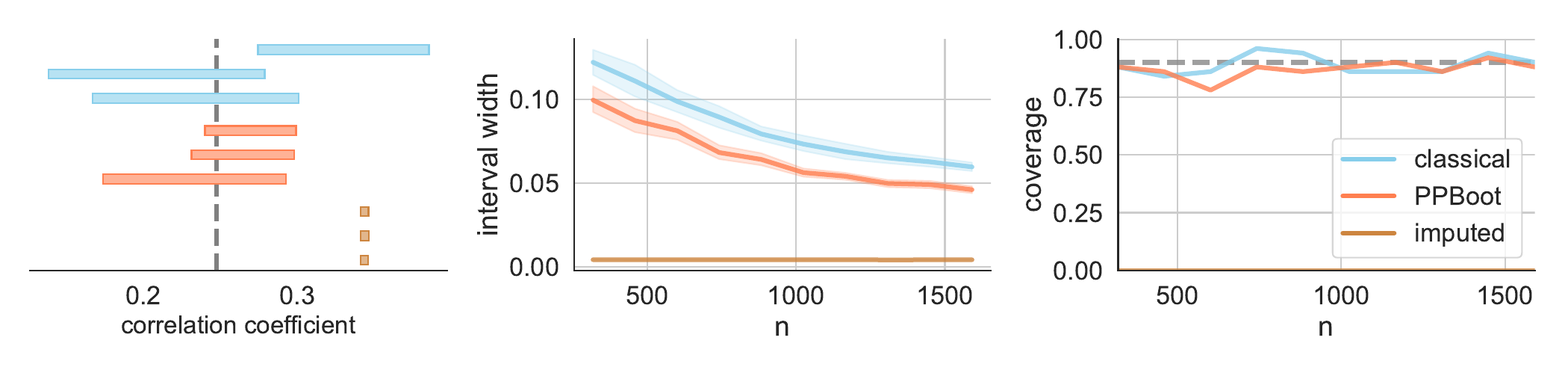}
\caption{Classical inference, \ppboot, and the imputed approach, applied to estimating the correlation coefficient between income and having health insurance in US census data.}
\label{fig:census_logistic_corr}
\end{figure}

To show that \ppboot~is applicable quite broadly, we also quantify the relationship between age and income and income and health insurance, respectively, using the Pearson correlation coefficient. Prior works on \ppi\texttt{(++)} do not study this problem, as the theory is not easy to apply to this estimand. To demonstrate that this is a nontrivial problem, we also form confidence intervals via the ``imputed'' approach, which naively treats the predictions as real data, and show that it severely undercovers the target. (For all previous applications, \citet{angelopoulos2023prediction} demonstrated the lack of validity of the imputed approach.)
We show the results in Figure \ref{fig:census_income_corr} and Figure \ref{fig:census_logistic_corr}, respectively. \texttt{PPBoot} yields smaller intervals than classical inference, while maintaining approximately correct coverage. In this application, classical inference uses the classical percentile bootstrap method. The imputed approach yields very small intervals and zero coverage.

\section{Extensions}
\label{sec:extensions}

We state two extensions of \texttt{PPBoot} that improve upon the basic method along different axes. The first one is a strict generalization of \texttt{PPBoot} that handles a version of ``power tuning'' \cite{angelopoulos2023ppipp}, leading to more powerful inferences than the basic method. The second one extends \texttt{PPBoot} to problems where the predictive model $f$ is not available a priori, the setting studied in \cite{zrnic2023cross}.

\subsection{Power-tuned PPBoot}

At a conceptual level, power tuning \cite{angelopoulos2023ppipp} is a way of choosing how much to rely on the machine-learning predictions so that their use never leads to wider intervals. In particular, power tuning should enable recovering the classical bootstrap when the predictions provide no signal about the outcome.

We define the power-tuned version of \texttt{PPBoot} by simply adding a multiplier $\lambda\in\R$ to the terms that use predictions:
\begin{equation}
\label{eq:lambdatuning}\theta_b^* = \lambda\cdot\hat \theta(\Xt^*,f(\Xt^*)) + \hat \theta(X^*,Y^*) - \lambda\cdot \hat \theta(X^*,f(X^*)).
\end{equation}
The parameter $\lambda$ determines the degree of reliance on the predictions:  $\lambda = 1$ recovers the basic \texttt{PPBoot}; $\lambda=0$ recovers the classical bootstrap. 

We will next show how to tune $\lambda$ from data as part of \texttt{PPBoot}. Before doing so, we derive the optimal $\lambda$ that the tuning procedure will aim to approximate. One reasonable goal is to pick $\lambda$ so that the variance of the bootstrap estimates $\Var(\theta^*_b)$ is minimized. A short calculation shows that the optimal tuning parameter for this criterion equals:
$$\lambda_{\mathrm{opt}} = \frac{\Cov\left(\hat\theta(X^*,f(X^*)), \hat \theta(X^*,Y^*)\right)}{\Var\left(\hat\theta(X^*,f(X^*))\right) + \Var\left(\hat\theta(\Xt^*, f(\Xt^*))\right)}.$$
To incorporate power tuning into \texttt{PPBoot}, we simply perform an initial bootstrap where we draw $(X^*,Y^*)$ and $\tilde X^*$---as in \texttt{PPBoot}---and compute
$$\hat\lambda_{\mathrm{opt}} = \frac{\widehat\Cov\left(\hat\theta(X^{*},f(X^{*})), \hat \theta(X^{*},Y^{*})\right)}{\widehat\Var\left(\hat\theta(X^{*},f(X^{*})) \right) + \widehat\Var\left(\hat\theta(\Xt^{*}, f(\Xt^{*}))\right)},$$
where the empirical covariance and variances are computed on the bootstrap draws of the data. After that, we proceed with \ppboot~as usual, the only difference being that we use the multiplier $\hat\lambda_{\mathrm{opt}}$, as in Eq.~\eqref{eq:lambdatuning}.

\begin{figure}[t]
\centering
\includegraphics[width=0.93\textwidth]{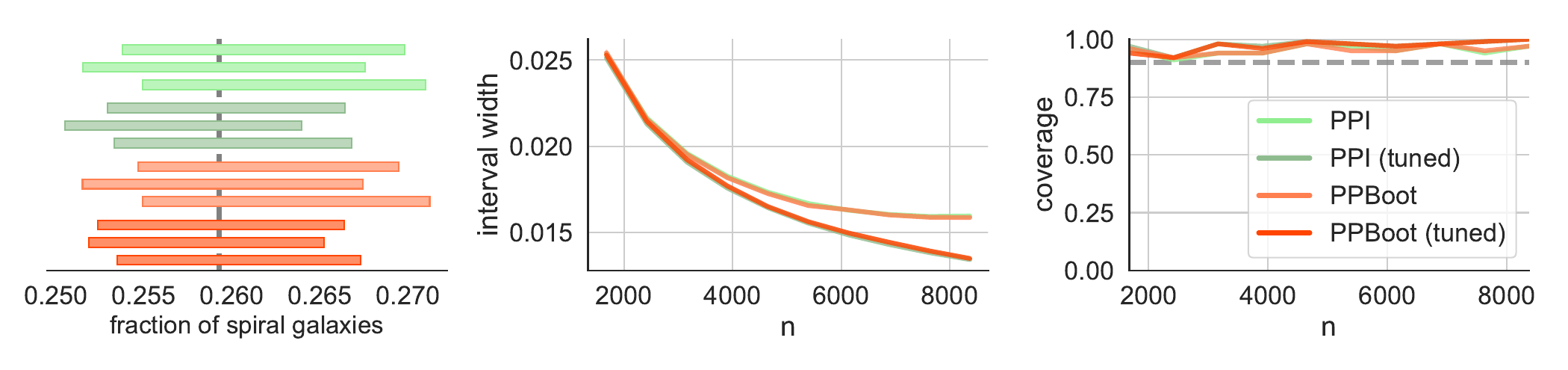}
\caption{\texttt{PPI}, \texttt{PPBoot}, and their tuned versions, applied to estimating the fraction of spiral galaxies from galaxy images.}
\label{fig:galaxy-tuned}
\end{figure}

\begin{figure}[t]
\centering
\includegraphics[width=0.93\textwidth]{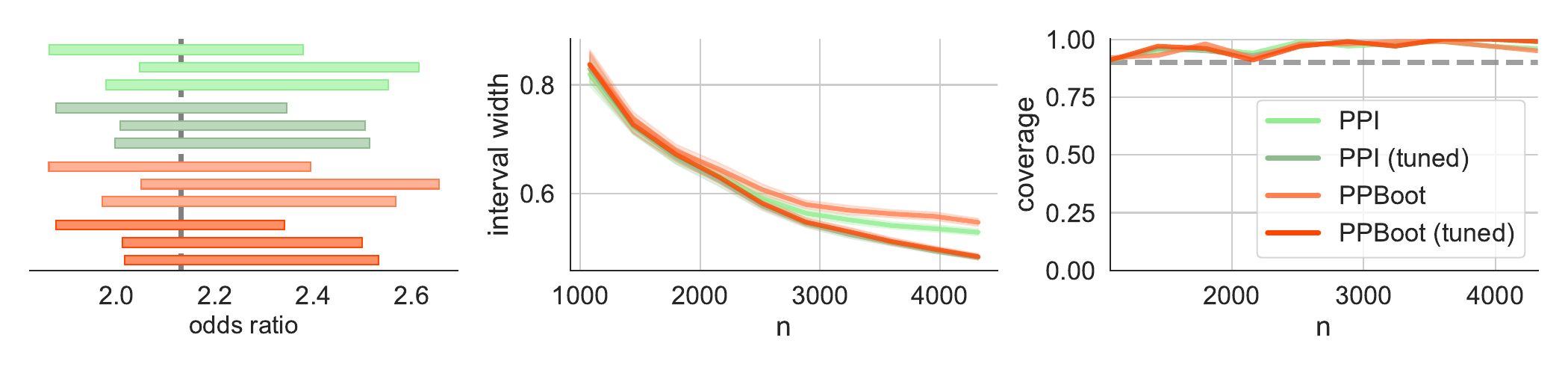}
\caption{\texttt{PPI}, \texttt{PPBoot}, and their tuned versions, applied to estimating the odds ratio between phosphorylation and protein disorder with AlphaFold predictions.}
\label{fig:alphafold-tuned}
\end{figure}

We evaluate the benefits of power tuning empirically. 
We revisit two applications: estimating the frequency of spiral galaxy via computer vision and estimating the odds ratio between phosphorylation and disorder via AlphaFold. The problem setup is the same as before, only here we additionally evaluate the power-tuned versions of \texttt{PPI} and \texttt{PPBoot}. We plot the results in Figure~\ref{fig:galaxy-tuned} and Figure~\ref{fig:alphafold-tuned}. As we saw before, \ppi~and \ppboot~yield intervals of similar size. Perhaps more surprisingly, the tuned versions also yield intervals of similar size, even though the tuning procedures are differently derived. For example, in the AlphaFold application, tuned \ppi~has two tuning parameters, while tuned \texttt{PPBoot} has only one. Also, as expected, the power-tuned procedures outperform their non-tuned counterparts.

\subsection{Cross-PPBoot}
Next, we extend \texttt{PPBoot} to problems where we do not have access to a pre-trained model~$f$. We therefore have the labeled data $(X,Y)$ and the unlabeled data $\tilde X$, but no $f$. This is the setting considered in \cite{zrnic2023cross}, and our solution will resemble theirs. It helps to think of \texttt{PPBoot} as an algorithm that takes $(X, Y, f(X))$ and $(\tilde X, f(\tilde X))$ as inputs:
\[\mathcal{C}^{\mathrm{\texttt{PPBoot}}} = \texttt{PPBoot}\left((X, Y, f(X)), (\tilde X, f(\tilde X))\right).\]
One obvious solution to not having $f$ is to simply split off a fraction of the labeled data and use it to train $f$. Once $f$ is trained, we use the remainder of the labeled data and the unlabeled data to apply \texttt{PPBoot}. Of course, this data-splitting baseline may be wasteful because we do not use all labeled data at our disposal for inference. Furthermore, we do not use all labeled data to train $f$ either.

To remedy this problem, we define \texttt{cross-PPBoot}, a method that leverages cross-fitting similarly to \texttt{cross-PPI} \cite{zrnic2023cross}. We partition the data into $K$ folds, $I_1,\dots,I_K$, $\cup_{j=1}^K I_j = [n]$. Typically, $K$ will be a constant such as $K=10$. Then, we train $K$ models, $ f^{(1)}, \dots,  f^{(K)}$, using any arbitrary learning algorithm, such that model $ f^{(j)}$ is trained on all labeled data except fold $I_j$. Finally, we apply \texttt{PPBoot}, but for every point $X_i$ in fold $I_j$, we use $f^{(j)}(X_i)$ as the corresponding prediction, and for every unlabeled point $\tilde X_i$, we use $\bar f(\tilde X_i) = \frac{1}{K} \sum_{j=1}^K  f^{(j)}(\tilde X_i)$ as the corresponding prediction. In other words, let $ f^{(1:K)}(X) = ( f^{(1)}(X_1), \dots,  f^{(K)}(X_n))$ be the vector of predictions corresponding to $X_1,\dots,X_n$, and let $\bar f(x) = \frac{1}{K} \sum_{j=1}^K  f^{(j)}(x)$ denote the average model. Then, we have 
\[\mathcal{C}^{\texttt{cross-PPBoot}} = \texttt{PPBoot}\left((X, Y,  f^{(1:K)}(X)), (\tilde X, \bar f(\tilde X))\right).\]
The cross-fitting prevents the predictions on the labeled data from overfitting to the training data.

We now show empirically that \texttt{cross-PPBoot} is a more powerful alternative to \texttt{PPBoot} with data splitting used to train a single model $f$. We also compare \texttt{cross-PPBoot} with \texttt{cross-PPI} \cite{zrnic2023cross}, since both are designed for settings where a pre-trained model is not available. \texttt{cross-PPI} leverages cross-fitting in a similar way as \texttt{cross-PPBoot}, however it computes confidence intervals based on a CLT rather than the bootstrap. As before, to avoid cherry-picking, we focus on the applications studied by \citet{zrnic2023cross}.

\begin{figure}[t]
\centering
\includegraphics[width=0.93\textwidth]{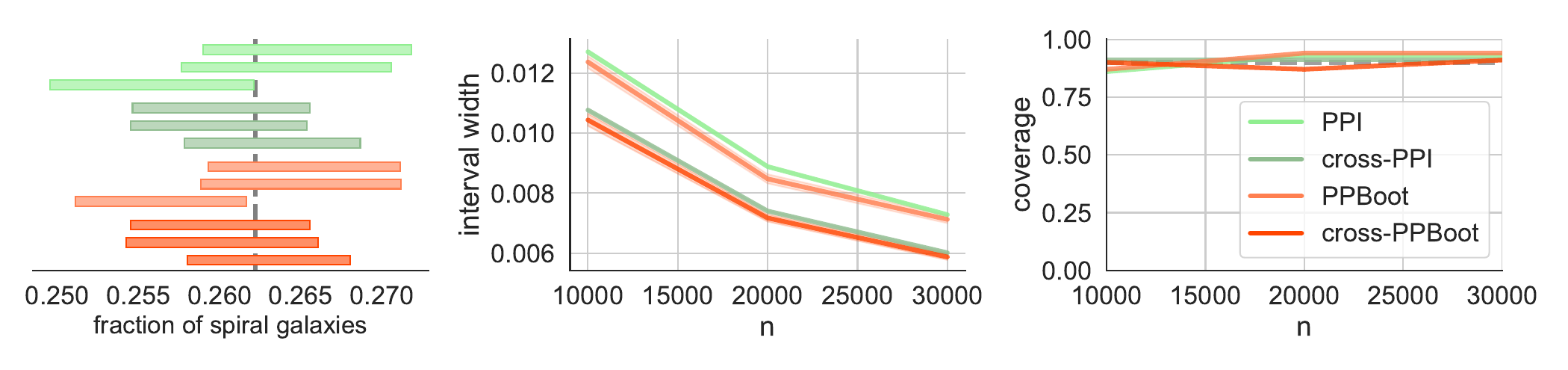}
\caption{\ppi, \crossppi, \ppboot, and \crossppboot, applied to estimating the fraction of spiral galaxies from galaxy images.}
\label{fig:galaxy-cross}
\end{figure}

\begin{figure}[t]
\centering
\includegraphics[width=0.93\textwidth]{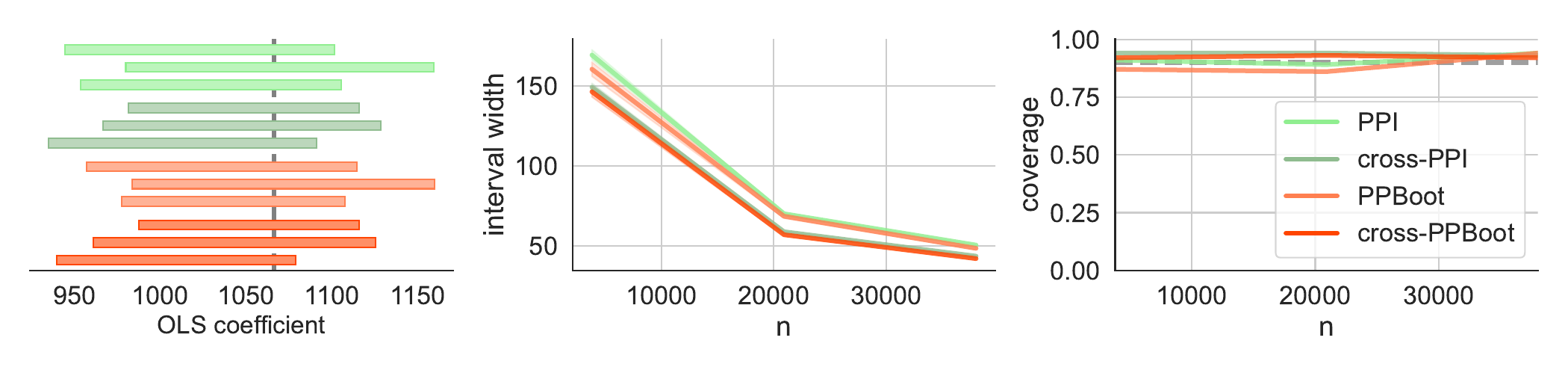}
\caption{\ppi, \crossppi, \ppboot, and \crossppboot, applied to estimating the relationship between age and income in US census data.}
\label{fig:census-cross}
\end{figure}

We consider the applications to spiral galaxy estimation from galaxy images and estimating the OLS coefficient between age and income in US census data. We use the same data and model-fitting strategy as in \cite{zrnic2023cross}. See Figure \ref{fig:galaxy-cross} and Figure \ref{fig:census-cross} for the results. In these two figures, \ppi~and \ppboot~refer to the data-splitting baseline, where we first train a model and then apply \ppi~and general \ppboot, respectively. Qualitatively the takeaways are similar to the takeaways from the power tuning experiments. The use of cross-fitting improves upon the basic versions of \ppi~and \ppboot, but again, somewhat surprisingly, \crossppi~and \crossppboot~lead to very similar intervals.

\section*{Acknowledgements}

T.Z. thanks Bradley Efron for many useful discussions that inspired this note.

\bibliographystyle{plainnat}
\bibliography{refs}

\end{document}